\documentclass[runningheads]{llncs}

\usepackage{eccv}

\usepackage[linesnumbered,ruled,vlined]{algorithm2e}

\usepackage{algpseudocodex}
\usepackage{amsmath}
\usepackage{amssymb}
\pdfoutput=1
\DeclareMathOperator{\E}{\mathbb{E}}

\usepackage{eccvabbrv}

\usepackage{graphicx}
\usepackage{booktabs}

\usepackage[accsupp]{axessibility}  

\usepackage[pagebackref,breaklinks,colorlinks,citecolor=eccvblue] {}

\usepackage{orcidlink}

\begin{document}

\title{Bayesian Active Learning for Semantic Segmentation} 


\author{Sima Didari\and
Wenjun Hu\and
Jae Oh Woo\and
Heng Hao\and
Hankyu Moon\and
Seungjai Min}

\authorrunning{S.~Didari et al.}

\institute{Samsung SDS Research America, San Jose, CA, USA\\
\email{hankyu.m@samsung.com}\\}

\maketitle

\begin{abstract}
  Fully supervised training of semantic segmentation models is costly and challenging because each pixel within an image needs to be labeled. Therefore, the sparse pixel-level annotation methods have been introduced to train models with a subset of pixels within each image.  We introduce a Bayesian active learning framework based on sparse pixel-level annotation that utilizes a pixel-level Bayesian uncertainty measure based on Balanced Entropy (BalEnt) \cite{jaeoh_iclr}. BalEnt captures the information between the models' predicted marginalized probability distribution and the pixel labels. BalEnt has linear scalability with a closed analytical form and can be calculated independently per pixel without relational computations with other pixels. We train our proposed active learning framework for Cityscapes, Camvid, ADE20K and VOC2012 benchmark datasets and show that it reaches supervised levels of mIoU using only a fraction of labeled pixels while outperforming the previous state-of-the-art active learning models with a large margin.
\end{abstract}

\section{Introduction}
\label{sec:intro}

\label{submission}
Best-performing semantic segmentation models are typically trained in a fully supervised setting depending on the availability of all-pixel annotation. This, by default, makes the data preparation process prohibitively costly and challenging to scale up. For example, approximately 1.5 hrs are required to annotate each image in the Cityscapes segmentation dataset \cite{Cityscapes}. To reduce the burden of annotations, Active Learning (AL) was suggested as an alternative training framework for reducing the annotation cost by selecting only the most informative samples for labeling \cite{caiSuperpixl, pixelpick, regimp}. 

One of the main challenges in AL is finding a representative subset of data with very little redundancy. In semantic segmentation, this subset can either be a collection of regions with predefined shapes such as boxes, superpixels, or a collection of individual pixels \cite{caiSuperpixl, pixelpick, metabox}. These regions can be annotated by simply `clicking' to select and assign their labels. However, an abundance of strongly correlated pixels exists in any given image carrying the same information. Therefore, selecting the most informative subset of pixels without any redundancy can significantly reduce the annotation cost \cite{pixelpick}. Motivated by this, we develop a highly data-efficient AL model utilizing sparse pixel annotation that reaches the level of supervised model accuracy within a few cycles using only a few pixels (\(\sim5 \)) per image per cycle. We extend the balanced entropy learning principle \cite{jaeoh_iclr} to a new pixel uncertainty measure in Bayesian AL and develop a Bayesian AL framework for semantic segmentation tasks. 
 
Numerous methods have been proposed toward optimal subset selection in the AL process, such as the deployment of specially tailored model architectures \cite{CoreGCN, VAAL, TA-VAAL, udd, uq_distil_segment, BayesSegNet, segeval} and loss functions \cite{learningloss}, core-set selection \cite{coreset}, utilizing sub-modular functions such as Determinantal Point Process \cite{dpp}\& SIMILAR \cite{similar}, using the model parameters as an acquisitions functions such as BADGE \cite{badge} and BAIT with Fisher information \cite{BAIT}. Additionally, low and high data regimes sampling strategy \cite{TypiClus} and optimal initial subset selection via pretext task learning \cite{initialseed_eccv, coldstart} have been studied. Uncertainty-based sampling with or without the introduction of some level of randomness to the sampling strategies \cite{nipsUq2020, bald, PowerBALD} and batch acquisition \cite{batchbald} methods have been studied as well. However, guaranteeing the exponentially-efficient subset selection method in the general AL framework is still an open problem \cite{zhu2022efficient, zhu2022active}. In this work we present a linearly scalable AL framework for semantic segmentation. 

Utilizing a Bayesian deep learning enabled modeling and measuring two main sources of uncertainty originating either from data (aleatoric) or models' parameters (epistemic) \cite{uq} for vision tasks. It was empirically demonstrated that these uncertainties play significant and mutually non-exclusive roles in the training of computer vision models \cite{gal_eps_aleo}. Therefore, it is crucial to keep track of these uncertainties throughout the AL process and design an optimal sampling strategy that takes into account the combined effects of various sources of uncertainty. Recently, the balanced entropy (BalEnt) learning principle for the Bayesian AL was introduced, and its linear scalability has been demonstrated \cite{jaeoh_iclr, wooISIT}. In BalEnt AL, instead of adding the samples to decrease the uncertainty, either models' or data \cite{bald,batchbald, PowerBALD}, samples are selected to balance the model's predictive power and the label uncertainty \cite{jaeoh_iclr}. A new uncertainty measure, called posterior uncertainty was introduced in addition to the aleatoric \& epistemic uncertainties to capture the model's uncertainty regarding its own predictions. Then, BalEnt uncertainty measure was defined as a combination of these three sources of uncertainty. Motivated by the superior performance of the BalEnt learning principle for image classification tasks, we present a Bayesian AL for semantic segmentation problems named BalEntAcq AL that utilizes a pixel-wise acquisition function based on the BalEnt uncertainty measure. We demonstrate that our approach has inherent diversification and exploration characteristics built-in, leading to the state-of-the-art performance even for cases with heavily under-representative categories using only a fraction of labels. Our contributions can be summarized as follows:

\textbf{1)} We adopt a new linearly scalable BalEnt measure \cite{jaeoh_iclr} for uncertainty estimation toward semantic segmentation AL scenarios that balances the pixels' epistemic and aleatoric uncertainties. We empirically verify that our acquisition function is capable of gradually selecting `harder' pixels (i.e., ranked by BalEntAcq) while decreasing the model's uncertainty,

\textbf{2)} We demonstrate that the pixel-based AL model with the BalEnt acquisition can select diverse samples, leading to the supervised level of accuracy across datasets,

\textbf{3)} We demonstrate that the proposed model reaches a desired accuracy level using a small fraction of labeled pixels across various backbones and datasets without the extra steps of generating superpixels, tailored loss functions, or tuning any hyper-parameter for the acquisition function.

\section{Related Works}
We can reduce the burden of annotation in semantic segmentation tasks from two angles: 1) developing models with the fewer annotation needs, such as weakly supervised, semi-supervised, and self-supervised. See section \ref{subsecweak}. 2) Using AL to label the most informative subset of the data. See sections \ref{subsecAL} and \ref{subsecALseg}. 

\subsection{Deep Learning Models and Data Dependency}\label{subsecweak}

Various modeling approaches exist to mitigate expensive labeling efforts, and are mainly grouped into weakly supervised \& semi-supervised, and self-supervised models \cite{Scribble, eyetrack, points, points2, webbase,boxsup, weaksup, boxdriven, xtreme, deepxtr,disloc, dilatedcnn, InterImg, semi_Al, bestpractice, RIPU} \cite{dataaug, pseudo, consistTrain, stego, DINO, diffAttSeg}. Recently, several zero-shot segmentation studies have emerged as a promising direction leveraging foundation models trained on large scaled data such as SAM \cite{SAM, SAM_Ultra} or vision-language models \cite{CLIP, Dens-CLIP} along with the existing unsupervised object localization \cite{FOUND} \& detection methods\cite{DINO, MoCoV2} \cite{CLIP-S, CLIP-DIY, CLIP-DINOiser, namedmask, ReCo}. However, it was shown that semantic segmentation models trained by AL had better performance compared to weakly and semi-supervised learning with far fewer annotations. Moreover,  constructing a model for zero-shot anything still remains a challenging task \cite{pixelpick, diffAttSeg, ACSeg}.

\subsection{AL for Computer Vision}\label{subsecAL}

AL approaches have adopted various subset selection strategies, such as uncertainty-based approaches in which an uncertainty measure is defined, and samples with the largest uncertainty are selected for labeling. Common measures of uncertainty include entropy \cite{bald}, margin sampling \cite{pixelpick}, and Bayesian-based measures such as Bayesian AL by disagreement (BALD) \cite{bald}, BatchBALD \cite{batchbald}, Causal-BALD \cite{Causal-BALD}, and PowerBALD \cite{PowerBALD}. BALD is defined as the mutual information between predictions and model parameters. In practice, selecting multiple samples at a single query is needed and if conducted naively, this batch selection is an intractable problem because the number of potential subsets grows exponentially with the data size. To resolve the issue, BatchBALD method has been introduced, which is a tractable approximation to calculate the mutual information between a batch of points and model parameters. PowerBALD was developed as a variant for BatchBALD to reduce its computational requirements using an importance sampling of tempered BALD scores. To overcome the problem of diversification facing uncertainty sampling methods, other acquisitions such as core-set \cite{coreset}, core-set hybrid with uncertainty sampling \cite{hybrid, hybrid_ensamble}, and loss-gradient-based \cite{badge} strategies were also developed. Reinforcement learning methods have been recently used as alternative acquisition functions and showed promising results, especially for imbalanced data \cite{ReAL,RLmeta}. We will compare the efficient BalEnt acquisition with these well-accepted acquisition functions in the experiments (see Section \ref{res}).

\subsection{AL for CNN-based Semantic Segmentation} \label{subsecALseg}
To reduce annotation cost, several AL semantic segmentation models were developed that have various levels of data selection granularity, such as region-based \cite{cereals, RLseg, metabox, KasarlaRegionBase} or pixel-based approaches \cite{pixelpick, regimp} versus the image-based selection \cite{deal,imbase_var}. For the region-based approaches the region's shape \& size such as irregular superpixels \cite{caiSuperpixl, seeds, viewal}, rectangle+polygon or rectangular boxes \cite{metabox, cereals, RLseg} shapes are further design choices. It was shown that labeling only few pixels per each region (either superpixel or box shapes) using a click-based annotation reduces the annotation cost \cite{caiSuperpixl}. Sparse pixel-based AL methods query only a fraction of pixels in each image per cycle and reduce the semantic segmentation annotation to a classification task \cite{pixelpick}.  Similarly, the proposed BalEntAcq AL framework also selects a few pixels in each image, which saves the labeling effort to the extreme.

To select the most informative subset of data either in the form of regions or pixels, various AL models have been developed. The main categories are uncertainty based \cite{pixelpick, viewal},  pseudo label based \cite{pseudoseg, pseudoAL} and deep reinforcement learning (RL) \cite{RLseg}. It was empirically shown that sparse-pixel AL frameworks with margin uncertainty-based sampling can reach acceptable accuracy while using substantially fewer labeled data\cite{pixelpick}. In margin sampling using the predicted softmax probabilities, samples having the smallest difference between their first and second predicted class probabilities are selected. However, because to boost the sampling diversification in the margin sampling some level of randomness is usually added. First, a random bigger subset of data is selected. Then the samples are ranked within this subset. The level of the randomness, i.e., the size of the initial subset is a hyper-parameter that needs to be tuned. However, BalEnt acquisition performs random sampling on each contour of BalEnt[$x$]=$c$ where \(c\geq0\). This random sampling on each contour line helps to select samples without any bias toward any specific class within the dataset \cite{jaeoh_iclr}. Thus, BalEnt acquisition is capable of selecting diverse samples automatically.

\section{Methods}
\begin{figure}[tb]
  \centering
  \includegraphics[width=0.9\textwidth]{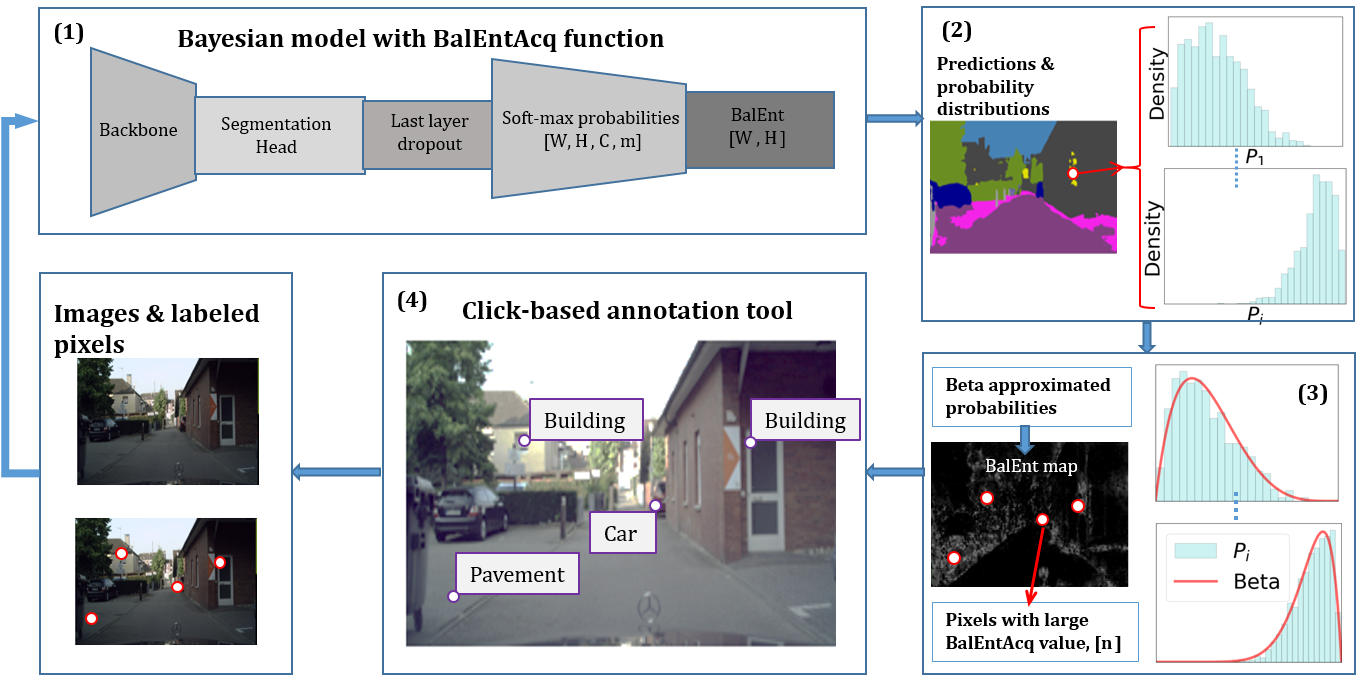}
\caption{Bayesian semantic segmentation model with last layer dropout and pixel-based loss. (2) Model's inference output \([W, H,C, m]\). Where \(C, m\) are the number of classes \& Monte Carlo samples. Per pixel probability distributions obtained from Monte Carlo model's forward pass with dropout are shown in histogram. (3) BalEnt acquisition function. Per-pixel probability distributions are approximated by a Beta distribution, shown in red curves. Per-pixel BalEnt values are calculated, and \(n\) pixels (here \(n\)=4) with the largest BalEnt values are selected, shown with red dots. (4) Click-based annotation tool. }
\label{figs:framework}
\vskip -0.2in
\end{figure}

In this section, we describe our BalEntAcq AL framework and its main components. The framework consists of four main components: a backbone-independent Bayesian deep learning model, a masked pixel-based loss function, a BalEnt acquisition function, and a click-based annotation tool. These four components are illustrated in Fig. \ref{figs:framework}. We briefly explain Bayesian deep learning models and the architecture used in this study in Section \ref{subsecBayesian}. The BalEnt uncertainty measure and acquisition function are introduced in Sections \ref{BetaApprx} \& \ref{subsecBalEnt}. Then, we describe the BalEntAcq AL model and its training pipeline in Section \ref{subsecFramework}. Other existing state-of-the-art acquisition functions are summarized in Section \ref{subsecAcq}.

\subsection{Bayesian Deep Learning Model}\label{subsecBayesian}
To quantify the uncertainty of neural networks, Bayesian deep learning models were introduced offering a mathematical framework to evaluate model and data uncertainty \cite{galdropout, laplacLastLayer}. In the BalEntAcq AL we use a dropout-based Bayesian deep learning where the distributions of class predictions are generated by Monte Carlo dropout sampling during the inference \cite{bald, neal}.  In our AL framework, we only add a single dropout layer to the last part (pixel classification) of the segmentation network architecture instead of applying dropouts throughout its backbone. This `last layer dropout' based Bayesian model is a simple and cost-effective architecture and its effectiveness for generating Gaussian predictive probability distributions has already been proven in the literature \cite{brossLastlayer, wilsonLastlayer, snoekLastLayer}. As in \cite{laplacLastLayer} we replace the last linear layer with a dropout layer with a ReLU activation function. In addition to its simplicity and cost-effectiveness, this proposed model architecture enables seamless integration of various semantic segmentation backbones to our framework.

\subsection{Beta Approximation for BalEntAcq AL Function} \label{BetaApprx}

In a dropout-based Bayesian deep learning model, distributions are generated by repeating Monte Carlo sampling of the model's forward pass. Therefore, these distributions are not presented in an analytically closed form. Not having the predicted distribution in a closed form impedes the utilization of various information-theoretic concepts and further analysis of the models' uncertainty. To resolve this issue, the softmax output of the dropout-based Bayesian deep learning network is approximated by a Beta probability distribution \cite{wooISIT, jaeoh_iclr}. \(\text{Beta}(\alpha\ , \beta \)) is a continuous probability distributions defined on [0, 1] interval with two positive parameters \(\alpha\) and \(\beta\) to control the shape of the distribution.

In the BalEntAcq AL, the last layer dropout Bayesian semantic segmentation model generates a pixel-wise probability distribution. By calculating the mean \& the variance of the histograms generated by the last layer dropout of the Bayesian neural network for each pixel, we estimate \(\alpha_i\) and \(\beta_i\), where \(i\) is indexed over \((1,\cdots, C)\) classes, and  fit a Beta distribution per each pixel class. A schematic of the histograms and their approximated Beta distributions (red line) are illustrated in Fig. \ref{figs:framework}.

\subsection{BalEnt Acquisition Function}\label{subsecBalEnt}

Framing the label acquisition in the AL process as a Bayesian learning procedure, BalEnt measure  takes into account the model's posterior uncertainty along with the epistemic and aleatoric uncertainties.  The posterior uncertainty is a measure for quantifying the certainty of the model in its predicted probabilities after the label acquisition. If a model is highly certain about its prediction after the label acquisition, then its expected posterior entropy should be low toward the negative direction.  

Marginalized Joint Entropy (MJEnt) \cite{jaeoh_iclr} introduced as the summation of these three different sources of uncertainties is defined in Equation (\ref{mje}): 

\begin{equation} \label{mje}
    \text{MJEnt}[x]:==  \sum_{i=1}^{C} {({\E} P{_i})h(P^+_i) } + H[x].
\end{equation} 
where \(H[x]\) is the Shannon entropy of the output over $P$ where $P=(P_1,\cdots,P_C)$ is the softmax probability prediction of each class, and \(h(\cdot)\) is the differential entropy. \(P^+_i\) is the conjugate posterior distribution of \(P_i\), with \(i\) indexed over the classes \(\{1,\cdots,C\}\).

The posterior uncertainty is defined as the expected differential entropy of the conjugate posterior distribution for all the classes, calculated by the first term in Equation (\ref{mje}), \(\sum_{i} {({\E} P{_i})h(P^+_i) }\). The posterior uncertainty is always non-positive and is maximized at 0 when each \(P^+_i\) is a uniform distribution. \(H[x]\) can be decomposed into two uncertainty measures as shown in Equation (\ref{entropy}): epistemic  uncertainty (the first term) and aleatoric uncertainty (the second term) \cite{jaeoh_iclr}.

\begin{equation}\label{entropy}
   H[x]= \ \mathfrak{I}(\hat{y}, \theta) + {\E_{\theta}}\left[ H({\hat{y}|\theta}) \right].
\end{equation}
where \(\mathfrak{I}\) is the generalized definition of mutual information between model's predictive outputs \(\hat{y}\) and its parameters \(\theta\) since it needs to accommodate both continuous and discrete domains \cite{mcfadden1965entropy, papangelou1978entropy, daley2007introduction, baccelli2016entropy}, and $H(\cdot|\theta)$ is the conditional Shannon entropy given the model parameters $\theta$.

BalEnt[\(x\)] defined in Equation (\ref{BalEnt}) is introduced as a re-scaled MJEnt by \(H[x] + \log 2\) \cite{jaeoh_iclr}. $\log$ base $e$ is used as a natural unit of information (nat) and $\log 2$ is added to the Shannon entropy with an additional entropy in an amount of $\log 2$ nats:
 \begin{equation}\label{BalEnt}
   \text{BalEnt}[x] := \frac{ \text{MJEnt}[x] } {\ H[x]+ \log 2}.
\end{equation}

 To calculate the posterior uncertainty we need to calculate the conjugate posterior distribution of \(P_i\), \(P^+_i\) for each image pixel $x$. Having the Beta distribution, then:  \(P^+_i\sim\text{Beta}(\alpha_i+1,\beta_i)\) which is the conjugate Beta posterior distribution of \(P_i\sim\text{Beta}(\alpha_i,\beta_i)\), then BalEnt [$x$] can be analytically calculated.

In the BalEnt learning principle, the samples are not selected to decrease the BalEnt uncertainty measure.  The samples are queried to fill the information gap and to push the BalEnt[\(x\)] toward its minimum which is \(0\) \cite{jaeoh_iclr}. The threshold $0$ is important since it is a natural choice of the information imbalance between the model and the label under some entropy amount of the floating number precision. And the positive value of BalEnt implies that there exists some information imbalance. Therefore, it has been shown that the sampling direction should be from BalEnt[\(x\)] $= 0$  toward its positively increasing direction. For the details, refer to Theorem 4.1 of \cite{jaeoh_iclr}. This direction is equivalent to choosing the entropy-increasing contours starting from  BalEnt[\(x\)] $= 0$. Because of the sampling on each contour, we expect that the acquired points would be naturally diversified. We will show empirical results on this diversification behavior in Section \ref{subsec:model_data_uncertainty}. Therefore, the reciprocal of BalEnt[\(x\)] is used \(\text{if BalEnt[$x$] $\ge $ $0$}\). This sampling strategy is formulated as the Balanced Entropy  Acquisition (BalEntAcq) function and defined as follows \cite{jaeoh_iclr}:
\begin{equation}\label{BalEnt_acq}
  {\text{BalEntAcq}[x]} :=
    \begin{cases}
      \ \text{BalEnt}[x]^{-1} & \text{if BalEnt[$x$] $\ge $ $0$}\\
      \ \text{BalEnt}[x] & \text{if BalEnt[$x$] $< $ $0$}.
    \end{cases}       
\end{equation}

\subsection{Proposed AL Framework for Segmentation}\label{subsecFramework}

The BalEntAcq AL framework is based on sparse pixel annotation. Instead of feeding the labels for the whole image or regions within the image (such as superpixels, boxes or polygons), we only train the model with a few labeled pixels per image at each AL cycle. To train the model with sparse pixel annotations we simply calculate the loss for the labeled pixels. Therefore, any existing semantic segmentation deep learning model can be trained using sparse pixel annotation. We use a cross-entropy loss for labeled pixels as defined in Equation(\ref{eq:xent}).
\begin{equation}\label{eq:xent}
     L(\theta^k, D_{L}^k) = - \frac{1}{|D_{L}^k|}\sum_{x\in D_{L}^k} { \sum_i^C {y_{i}\cdot \log ({\E} P{_i})} } .
\end{equation} 
where \(i\), \(D_{L}\), \(k\), \(x\), \({\E} P{_i}\) and \(y_{i}\) are the index over \((1,\cdots, C)\) classes, labeled pixel data, AL cycle, pixel coordinates, model prediction for \(i\)-th class, and ground truth label for the \(i\)-th class, respectively \cite{pixelpick}. Note that $\theta^k$ and $D_L^k$ mean model parameters and labeled data at the cycle $k$.

The BalEntAcq AL model training process is explained in Algorithm 1. This iterative process begins with randomly sampled \(n\) pixels per image for labeling from a given unlabelled dataset  \(D_{pool}\). The Bayesian semantic segmentation model is trained with this initial labeled data. The model's pixel-wise prediction distributions for \(D_{pool}\) are passed to the acquisition function to estimate their per-pixel uncertainties and select \(n\) pixels having the largest BalEntAcq values, to be labeled for the next cycle. The segmentation model is retrained on the expanded labeled pixels, the uncertainty scores of the remaining unlabeled pixels are calculated, and the new sets of pixels are selected for annotation. These steps are repeated until the annotation budget (a predefined number of cycles) is exhausted or the accuracy of the model reaches a satisfactory level. In this study, we feed already prepared ground truth labels incrementally to simulate such an annotation process. Since the BalEntAcq is a stand-alone measure without any need for further information from neighboring pixels, pixel-wise BalEnt calculations can be run in parallel to reduce the computation time \cite{jaeoh_iclr}. 

\begin{algorithm}[tb]
   \caption{BalEntAcq AL framework algorithm}\label{alg:example}

    {Pick the number of pixels \(n\), total AL iterations \(K_{tot}\), \(k\) := 0, \(D_{L} := \emptyset \).}\\
    {Initialize the model with pre-trained backbone weights.} \\
   {Randomly sample \(n\) pixels per image from  \(D_{pool}\).}\\
\While{\(k < K_{tot}\)}
   {Add selected pixel labels to \(D_{L}\).  \\
   Train the model with \(D_{L}\) and loss from Equation (\ref{eq:xent}).\\
   Run inference \(m\) times for \(D_{pool}\), save pixels probability distributions.\\
    Calculate BalEntAcq for each pool pixel/image using Equation (\ref{BalEnt_acq}).\\
   Remove the selected \(n\) pixels/image from the \(D_{pool}\).\\
    \(k:= k+1\)}

\end{algorithm}
\vskip -0.2in

\begin{figure}[tb]
  \centering
\includegraphics[width=0.85\textwidth]{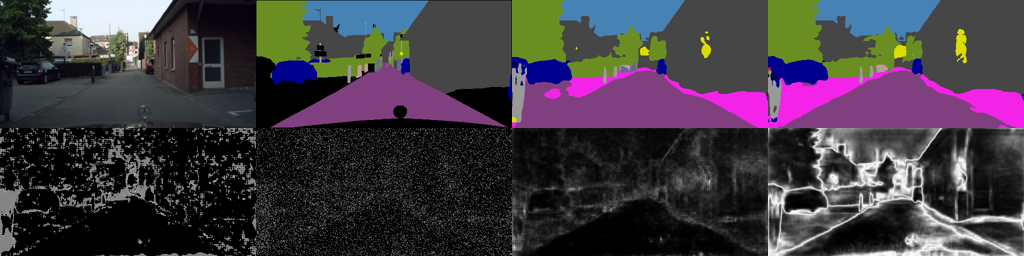}
\caption{Top row, left to right: Input image, its ground truth, supervised training prediction, BalEntAcq AL prediction, DeepLab,  \(n\)=5. Bottom row:  Uncertainty maps from left to right: BalEnt, pBALD, BALD, \& \(P_{marg}\). Brighter intensity represents higher uncertainty values. }
\label{figs:ucmap}
\vskip -0.1in
\end{figure}

\subsection{Other Acquisition Functions}\label{subsecAcq}
In this section, we summarize well-accepted uncertainty measures for AL and will compare their performance versus BalEnt later on:

\textbf{1)} Random: \(\text{Rand}[x] := U\) where  \(U\) is a uniform distribution over $[0,1]$, independent of  \(x\). The random acquisition function assigns a random uniform value to each pixel. 

\textbf{2)} Entropy \cite{shannon1948mathematical}:  \( H[x] :=  - \sum_i {\mathbb{E} P_i \cdot \log \mathbb{E}P_i}\)  where \(H[x]\) is the Shannon entropy of the predictive output over the probability $P=\left(P_1\cdots,P_C\right)$. 

\textbf{3)} Margin Sampling: \( P_{marg} [x]:= \max \mathbb{E}P_i - \max_2 \mathbb{E}P_i \) is defined as the difference between the first and the second most probable labels' probabilities, where $\max_2$ means the second-largest component. We select pixels having as small $P_{marg}$ as possible. However, it was shown that the pure \(P_{marg}\) is not effective to diversify the samples. Thus, some level of randomness was introduced to this sampling method \cite{pixelpick}. In this work, we first select a random pool of data $10$ times bigger than the sampling budget. Then, within this pool samples are selected using \(P_{marg}\).

\textbf{4)} BALD \cite{lindley1956measure, houlsby2011bayesian,bald}: \(\text{BALD}[x]:=\mathfrak{I}(\theta,  \hat{y})\). BALD leverages the epistemic uncertainty of the model to the output, which is the mutual information between the model parameters $\theta$ and the predictive output \(\hat{y}\).

\textbf{5)} PowerBALD \cite{farquhar_statistical_2020, PowerBALD}: \( \text{pBALD}[x]:=  \log \text{BALD}[x] + Z \)  where PowerBALD is a randomized BALD by adding a power-law distribution, i.e., $Z$ is an independently generated random number from Gumbel distribution with exponent $\gamma>0$. The randomization is for diversifying the selections using a weighted sampling strategy by putting weights onto the high BALD values proportional to $\text{BALD}[x]^\gamma$. We note that PowerBALD is a linearly scalable version of BALD with diversification as opposed to BatchBALD \cite{batchbald}, which is not computationally feasible for semantic segmentation tasks. In this experiment, we use $\gamma=1$ as a default choice suggested by PowerBALD \cite{PowerBALD}.

\section{Experiments}
We introduce the datasets used in the experiments, the network architectures, training details, \& the main results of BalEntAcq AL in Sections \ref{data}, \ref{expsetting}, \& \ref{res}. 

\subsection{Datasets}\label{data}
We use four datasets, PASCAL VOC 2012 \cite{voc}, CamVid \cite{CamVid}, Cityscapes \cite{Cityscapes} and ADE20K \cite{ADE20K}. VOC2012 has 20 categories and $1464$, $1449$, and $1456$ images for training, validation and testing.  Since the images of VOC2012  have different sizes, during training we resize to $496$×$496$. CamVid  is an urban scene segmentation dataset with $11$ categories, consisting of $367$, $101$, and $233$ images of $480$×$360$ resolution for training, validation, and testing, respectively. Cityscapes have high-resolution images of $2048$×$1024$ with $19$ classes and are collected for autonomous driving. It consists of $2975$ \& $500$ training  and validation images, respectively. To speed up the training,  the dataset is resized to $512$×$256$ pixels \cite{pixelpick}. ADE20K consists of a  diverse set of visual concepts with $150$ categories, with $20210$, $2000$, and $3000$ images that we resized to $320$×$320$ resolution for training, validation, and testing, respectively.

\begin{figure}[htbp]
\centering
\includegraphics[width=0.9\textwidth]{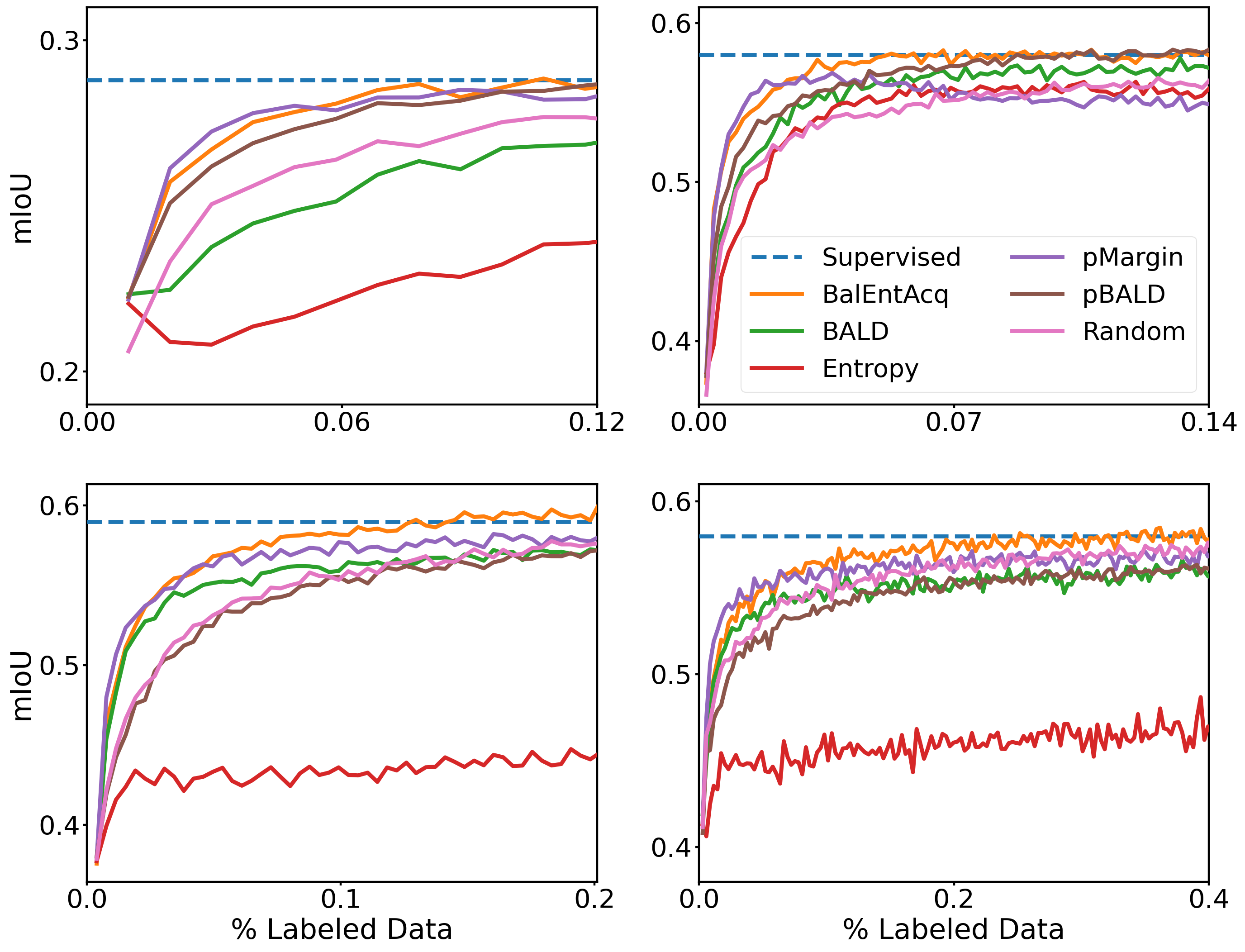}
\caption{The BalEntAcq AL comparison to the existing acquisition functions, ADE20K, VOC2012, Cityscapes, \& CamVid (from top to right). DeepLabV3+ MobileNetv2, $n$=10 for ADE20K, $n$=5 for the rest of datasets.}
\label{figs:deeplab_res}
\vskip -0.1in
\end{figure}

\begin{table}[tb]
\caption{mIoU for various acquisition functions when BalEntAcq AL reaches (or the closest to) supervised mIoU, DeepLabV3+ MobileNetv2, $n$=10 for ADE20K, $n$=5 for the rest of datasets.}
\label{dataEff}
\begin{small}
\begin{sc}
\resizebox{0.98\textwidth}{!}
{
\begin{tabular}{llllllll}
\toprule
Datasets & BalEntAcq & BALD & pBALD & pMargin &Entropy  & Random & Supervised\\
\midrule
Cityscapes &  $\bf{0.591(\pm 0.003)}$ &  $0.564(\pm 0.002)$ &  $0.561(\pm 0.005)$ &  $0.574(\pm 0.002)$ &  $0.432(\pm 0.011)$ &  $0.566(\pm 0.005)$ & $0.59$ \\
CamVid  &  $\bf{0.581(\pm 0.006)}$ &  $0.556(\pm 0.002)$ &  $0.556(\pm 0.001)$ &  $0.566(\pm 0.006)$ &  $0.463(\pm 0.003)$ &  $0.568(\pm 0.006)$ &    $0.58$ \\
VOC2012 &  $\bf{0.581(\pm 0.004)}$ &  $0.565(\pm 0.003)$ &  $0.568(\pm 0.002)$ &  $0.560(\pm 0.003)$ &  $0.552(\pm 0.002)$ &  $0.546(\pm 0.008)$ &    $0.58$ \\
ADE20K &  $\bf{0.288(\pm 0.001)}$ &  $0.268(\pm 0.001)$ &  $0.285(\pm 0.001)$ &  $0.282(\pm 0.003)$ &  $0.238(\pm 0.002)$ &  $0.277(\pm 0.003)$ &    $0.29$ \\
\bottomrule
\end{tabular}}
\end{sc}
\end{small}
\end{table}


\begin{figure}[tb]
\begin{center}
\centerline{\includegraphics[width=\textwidth]{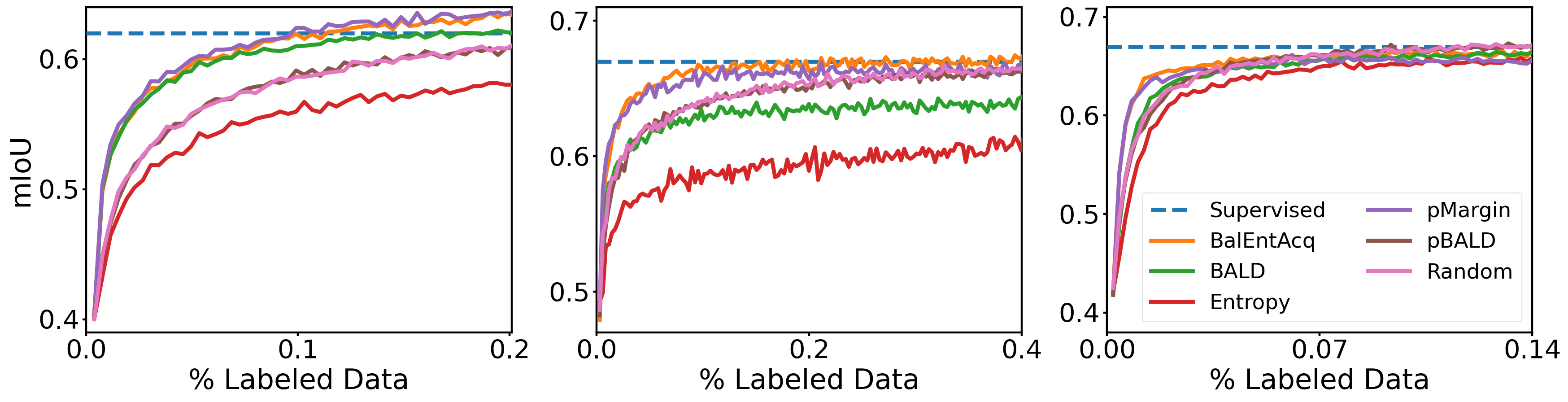}}
\caption{The BAlEntAcq AL comparison to existing acquisition functions, Cityscapes, CamVid \& VOC2012 (from left to right), FPN ResNet50, \(n\)=5.}
\label{figs:ResNet50_res}
\end{center}
\vskip -0.3in
\end{figure}

\begin{table}[tb]
\caption{mIoU for various acquisition functions when BalEntAcq AL reaches (or the closest to) supervised mIoU, FPN ResNet50, \(n\)=5.}
\vskip -0.1in
\label{dataEff_ResNet50}
\begin{small}
\begin{sc}
\resizebox{0.98\textwidth}{!}
{
\begin{tabular}{llllllll}
\toprule
Datasets & BalEntAcq & BALD & pBALD & pMargin &Entropy  & Random & Supervised\\
\midrule
Cityscapes  &  $0.621(\pm 0.002)$ &  $0.615(\pm 0.002)$ &  $0.593(\pm 0.004)$ &  $\bf{0.627(\pm 0.002)}$ &  $0.566(\pm 0.001)$ &  $0.509(\pm 0.003)$ &    $0.62$ \\
CamVid  &  $\bf{0.671(\pm 0.004)}$ &  $0.635(\pm 0.005)$ &  $0.651(\pm 0.009)$ &  $0.667(\pm 0.006)$ &  $0.598(\pm 0.011)$ &  $0.653(\pm 0.003)$ &    $0.67$ \\
VOC2012 & $0.664 (\pm 0.001)$ &  $0.657 (\pm 0.003)$ &   $\bf{0.672 (\pm 0.001)}$ &  $0.659(\pm 0.002)$ &  $0.653(\pm 0.001)$ &  $0.667(\pm 0.002)$ &   $0.67$ \\

\bottomrule
\end{tabular}}
\end{sc}
\end{small}
\end{table}



\subsection{Experimental Settings} \label{expsetting}
 We choose two architectures (1) MobileNetv2 \cite{MobileNetv2} implemented in DeepLabv3+ \cite{Deeplab} and (2) Feature Pyramid Network (FPN) \cite{FPN} with a dilated ResNet \cite{ResNet}, which replaces the last two residual blocks with atrous convolutions \cite{atrous}. We use the cross-entropy loss only on the labeled pixels and train the model for $100$ epochs. We use a learning rate decay and learning rate scheduler as in \cite{pixelpick}. We apply random scaling between [$0.5$, $2.0$], random horizontal flipping, color jittering, and Gaussian blurring as suggested in \cite{dataaug}. There are three main input parameters: number of Monte Carlo forward passes (\(m\)), dropout ratio (\(r\)), and the number of pixels per image queried for labeling  (\(n\)). \(m\) and \(r\) are fixed at $20$ \& $0.2$, respectively. We report the intersection over union (mIoU) averaged over three repetitions with different random seeds for all the experiments. 

\subsection{Results}\label{res}
To study the effectiveness of the BalEntAcq AL model we compare its performance to the AL models with the existing uncertainty-based acquisition functions. In Fig. \ref{figs:ucmap} a sample of uncertainty maps for BalEntAcq, pBALD, BALD, and margin sampling methods are shown. The brighter the pixel's intensity, the higher its uncertainty value is. The BalEnt values have a wide range within an image. So, for illustration purposes, they are scaled with the normal Cumulative Distribution Function (CDF) as \(norm.cdf(uc)/100\) where \(uc\) represents BalEnt uncertainty per pixel.  As can be observed in the figure, large BalEnt values are scattered throughout the image and are not dominantly concentrated around object boundaries in contrast to the \(P_{marg}\) and BALD values. A large fraction of pixels have high pBALD values which is not an optimal scenario if we want to select only a few most informative pixels with large uncertainty values for labeling. We postulate that the dispersion of BalEnt values results in higher diversification in sample selection and better model performance. 
\subsubsection{Data efficiency}
The BalEntAcq AL reaches the supervised accuracy given the specific backbone used using merely $0.13$\% of labeled pixels for Cityscapes, $0.25$\% for CamVid, $0.05$\%  for VOC2012 and $0.11$\% for ADE20K datasets when using DeepLabV3+ MobileNetv2, as shown in Fig. \ref{figs:deeplab_res}. We also tested BalEntAcq AL with FPN ResNet50 backbone for Cityscape, CamVid \& VOC2012 datasets. As shown in Fig. \ref{figs:ResNet50_res} BalEntAcq AL reaches the supervised level accuracy with only $0.11$\%,  $0.18$\%  and $0.09$\% of labeled pixels, respectively. As depicted in Tables \ref{dataEff} \& \ref{dataEff_ResNet50}, BalEntAcq AL either outperforms the other existing acquisition functions or is close to the best performing acquisition functions (such as pBALD or pMargin) when it approaches the supervised mIoU values. 

\begin{figure}[tb]
\begin{center}
\centerline{\includegraphics[width=\textwidth]{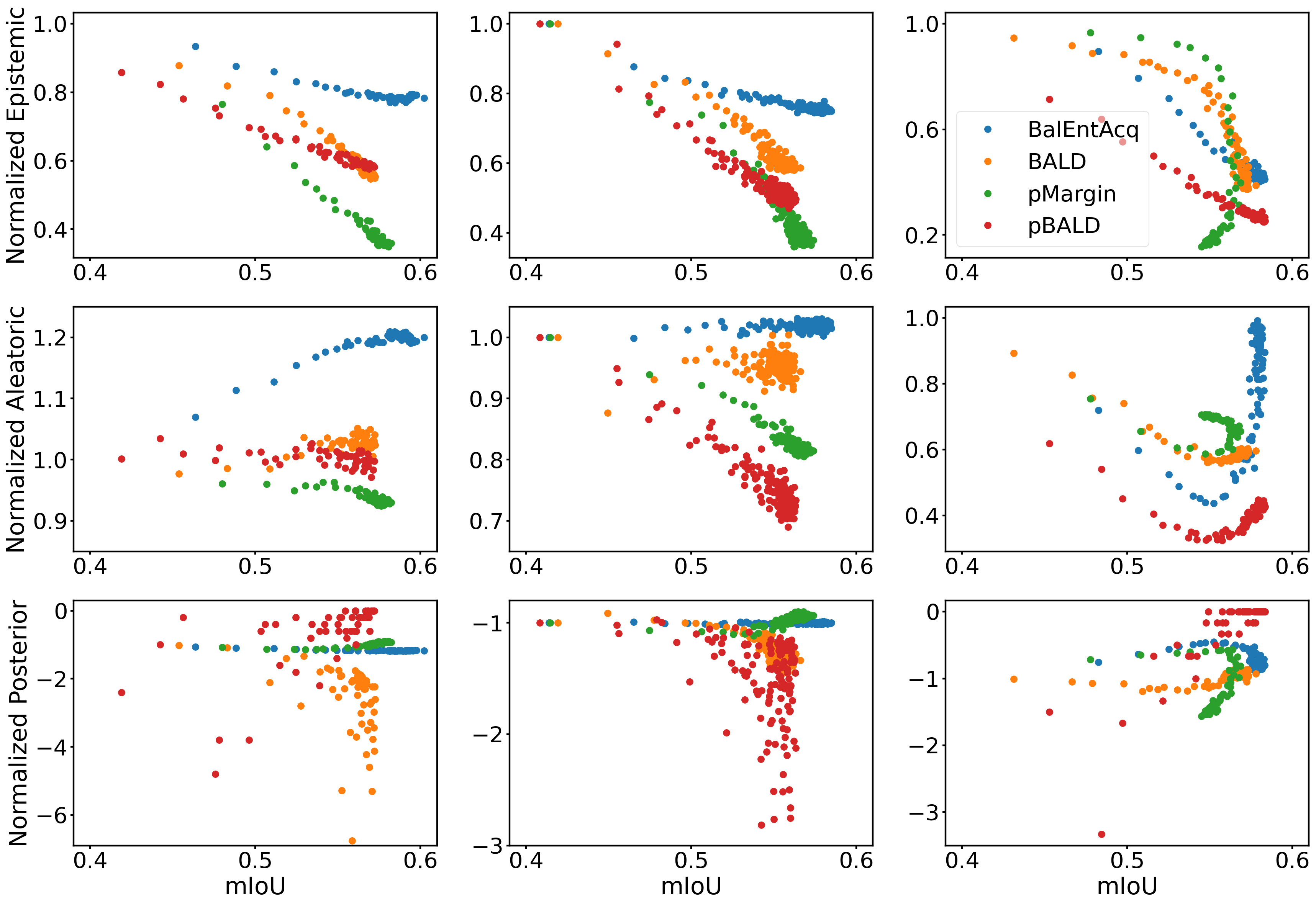}}
\caption{Normalized epistemic, aleatoric \& posterior uncertainties with their values at the first cycle versus the validation dataset mIoU for Cityscapes, CamVid \& VOC2012 (from left to right), DeepLabV3+ MobileNetv2, $n$=$5$, each point represents an AL cycle. }
\label{figs:ucn}
\end{center}
\vskip -0.5 in
\end{figure}


\begin{figure}[tb]
\begin{center}
\centerline{\includegraphics[width=\textwidth]{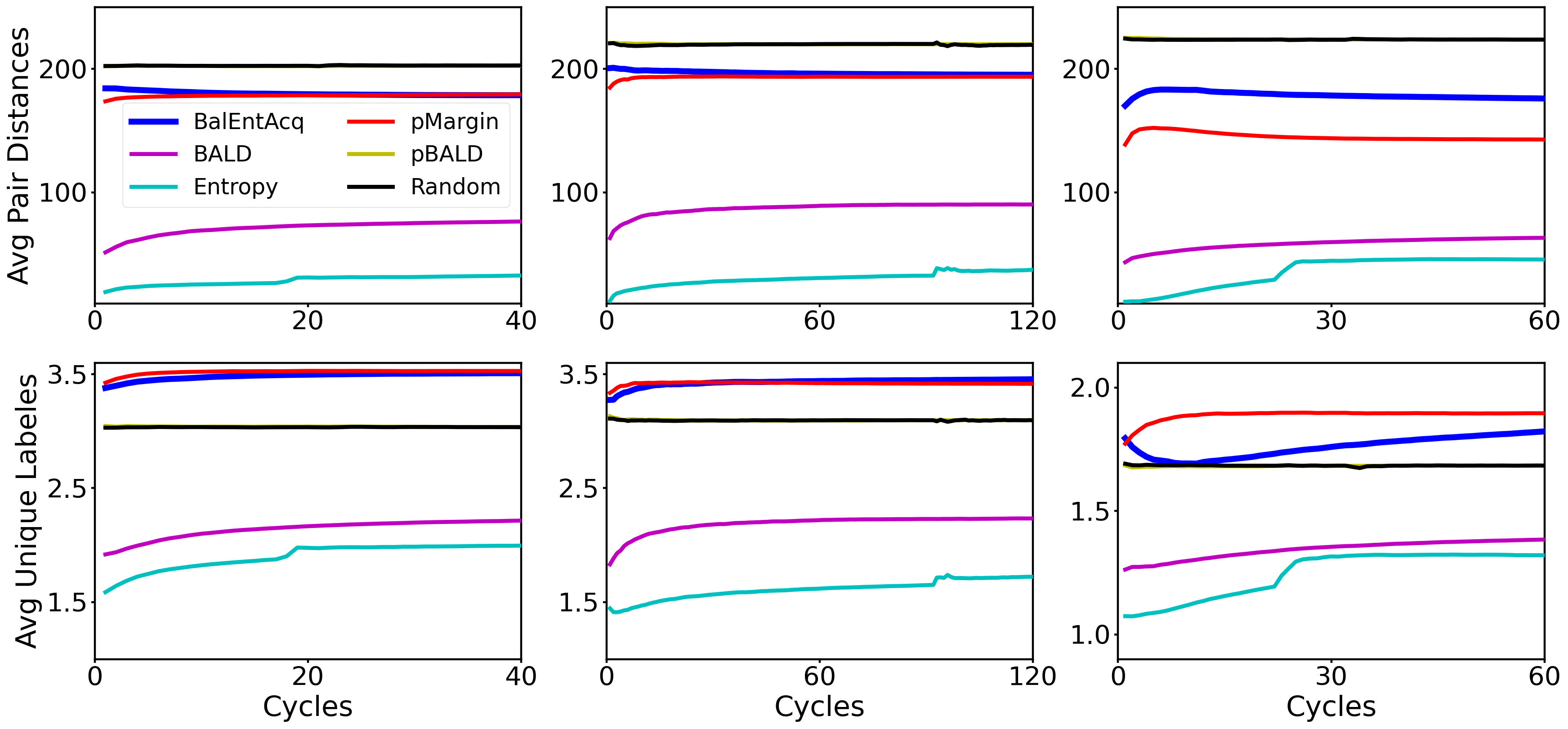}}
\caption{Pair-wise distance of pixels \& average unique labels queried at each cycle for Cityscapes, CamVid \& VOC2012 dataset, (from left to right), DeepLabV3+ MobileNetv2, $n$=$5$. }
\label{figs:diversity}
\end{center}
\vskip -0.3in
\end{figure}


\begin{figure}[tb]
\begin{center}
\centerline{\includegraphics[width=\textwidth]{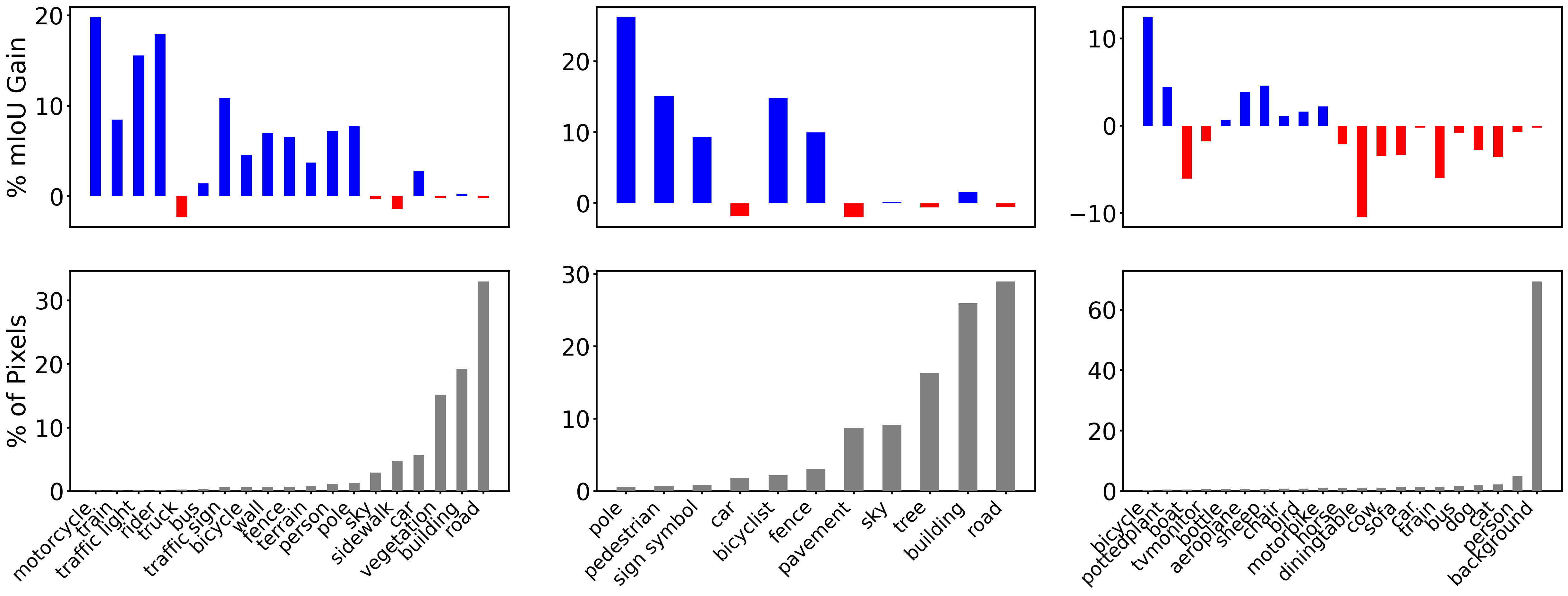}}
\caption{Per-class mIoU gain for Cityscapes, CamVid \& VOC2012 (from left to right) datasets, DeepLabV3+ MobileNetv2 for BalEntAcq sampling, \(n\)=5.}
\label{figs:imb_deeplab}
\end{center}
\vskip -0.45in
\end{figure}


\subsubsection{Model and data uncertainty in AL}\label{subsec:model_data_uncertainty}
To have a better understanding of the individual contribution of the epistemic, aleatoric, and posterior uncertainty measures in the BalEntAcq AL model, we track their changes throughout the AL training cycles. Then we compare these measures under standard Bayesian AL uncertainty sampling strategies, BALD, \& pBALD and \(P_{marg}\). For the purposes of illustration, we normalize the uncertainty measures by their values at the first cycle and plot the averaged and normalized uncertainty measures for the queried pixels versus the validation dataset's mIoU at each cycle.  As shown in Fig. \ref{figs:ucn}, all the sampling strategies reduce the epistemic uncertainty throughout the AL training process; as the labeled data volume grows the model accuracy increases. Therefore its epistemic uncertainty naturally decreases. However, for BalEntAcq sampling there is no explicit constrain on the aleatoric uncertainty, and as shown in Fig. \ref{figs:ucn} the aleatoric uncertainty gradually changes through the low to high data regimes in the AL process, enabling optimal sampling to decrease models' uncertainty. Also, with BalEntAcq no extra randomness has been injected, leading to more robustness in the posterior uncertainty through the AL process.  We hypothesize that capturing the interplay between these three uncertainties is the key to the superior performance of BalEntAcq sampling.

To further quantify and validate the diversification and exploration features of the sampling methods, we use two metrics: average pair distance (the average of the pairwise Euclidean distance of pixels queried) and average unique labels (the average of the unique pixel labels that are queried at each cycle). The average pair distance is an indicator of the dispersion of the selected pixels within an image and the average unique labels represent the diversification capability of an acquisition function. As shown in Fig. \ref{figs:diversity}, BalEntAcq sampling selects pixels that are scattered throughout the image with label diversity as well. Both average pair distance \& average unique labels are larger in BAlEntAcq compared to BALD and Entropy sampling and similar to random, \(P_{marg}\) and pBALD. The latter two acquisition functions inject randomness explicitly (a predefined hyper parameter) to increase their sampling diversification, while BalEntAcq sampling can inherently select a set of diverse samples without the introduction of any randomness or the need to tune the percentage of randomness. 

Due to the inherent diversification and exploration capabilities of BalEntAcq it performs well for underrepresented classes as well, leading to mIoU gain compare to the supervised training for the tail classes, as shown in Fig. \ref{figs:imb_deeplab}. 

\section{Conclusion}

We have introduced a new AL method, BalEntAcq AL, for semantic segmentation that leverages the BalEnt uncertainty measure embedded in a Bayesian Deep Learning framework.
We have proposed pixel selection as a data-efficient active learning strategy. It allows us to acquire labels for only the most informative pixels from each image, leading to substantial savings in labeling efforts. While we have shown results on well-known datasets using popular DeepLab \& FPN models, our method is agnostic to data formats and model backbones, and can easily be applied to a wide range of scenarios. We have shown the effectiveness of pixel-based BalEntAcq AL by evaluating its accuracy as a function of the percentage of labeled pixels and demonstrated its advantage over other state-of-the-art approaches on the same benchmark datasets.
We also have shown that BalEntAcq AL provides inherent diversification and exploration properties that result in performance on par with supervised training  even for imbalanced classes, without utilizing weighted loss or fine-tuning model architectures. Moreover, we demonstrated that the superior performance of our BalEntAcq AL is due to its combination of model (epistemic), data (aleatoric), and model bias (posterior) uncertainties concurrently throughout the AL model training.

\bibliographystyle{splncs04}
\bibliography{main}
\end{document}